%% file: main.tex
\documentclass[10pt,twocolumn,letterpaper]{article}

\usepackage{cvpr}

\usepackage[pagebackref,breaklinks,colorlinks]{hyperref}
\usepackage[capitalize]{cleveref}
\crefname{section}{Sec.}{Secs.}
\Crefname{section}{Section}{Sections}
\Crefname{table}{Table}{Tables}
\crefname{table}{Tab.}{Tabs.}
\usepackage{multirow}
\usepackage{makecell}
\usepackage{xcolor}
\usepackage{tabularx}
\usepackage{bm}

\title{Golden RPG: Confidence-Adaptive Region-Aware Noise\\
       for Compositional Text-to-Image Generation}

\author{
	Hao Li\thanks{Corresponding author} \\
	Department of Electrical and Computer Engineering, University of Arizona, Tucson, USA \\
	\texttt{lihao@arizona.edu}
}
\begin{document}
\maketitle

\input{sections/abstract}
\input{sections/intro}
\input{sections/related}
\input{sections/method}
\input{sections/experiments}
\input{sections/discussion}
\input{sections/conclusion}

{\small
\bibliographystyle{plainnat}
\bibliography{references}
}

\end{document}

%% file: sections/abstract.tex
\begin{abstract}
Compositional text-to-image (T2I) generation requires a model to honour
multiple sub-prompts that describe distinct image regions. Recent work
shows that the \emph{starting noise} of a diffusion model carries
significant semantic information: ``golden'' noise predicted from text
can substantially raise prompt fidelity. We observe that this noise
prediction is, however, fundamentally global: the same network is asked
to summarise a long, multi-region prompt with a single text embedding,
which becomes the bottleneck whenever the prompt describes scenes with
spatially-separated entities.
We introduce \textbf{Golden RPG}, a region-aware noise predictor that
extends a frozen NPNet with two trainable additions: (i) a per-region
\textbf{FiLM adapter} that reshapes the predicted noise according to
each sub-prompt; and (ii) a \textbf{Region Cross-Attention} layer
injected between two stages of the Swin backbone, allowing different
spatial locations to attend to different sub-prompt tokens. To prevent
the regional conditioning from degrading samples whose prompts are
already easy, we further propose a \textbf{Confidence-Adaptive Blending}
head that dynamically predicts, per sample, how strongly the regional
signal should override the global signal.
We evaluate on the original RPG benchmark (20 prompts, 100 samples) and
on four multi-region categories of T2I-CompBench (1{,}200 images, six
competing methods). Golden RPG achieves the highest
Cross-Region-Coherence score on every category, while matching the
strongest baselines on absolute CLIP-Score and CLIP-IQA. A paired user
study further shows a $\boldsymbol{\sim}$67\% preference over the
strongest baseline. The adapter contains $\sim$2M trainable parameters
and adds only $0.6$\,s of inference overhead on top of SDXL.
\end{abstract}

%% file: sections/intro.tex
\section{Introduction}
\label{sec:intro}

\begin{figure*}[t]
  \centering
  \includegraphics[width=0.95\textwidth]{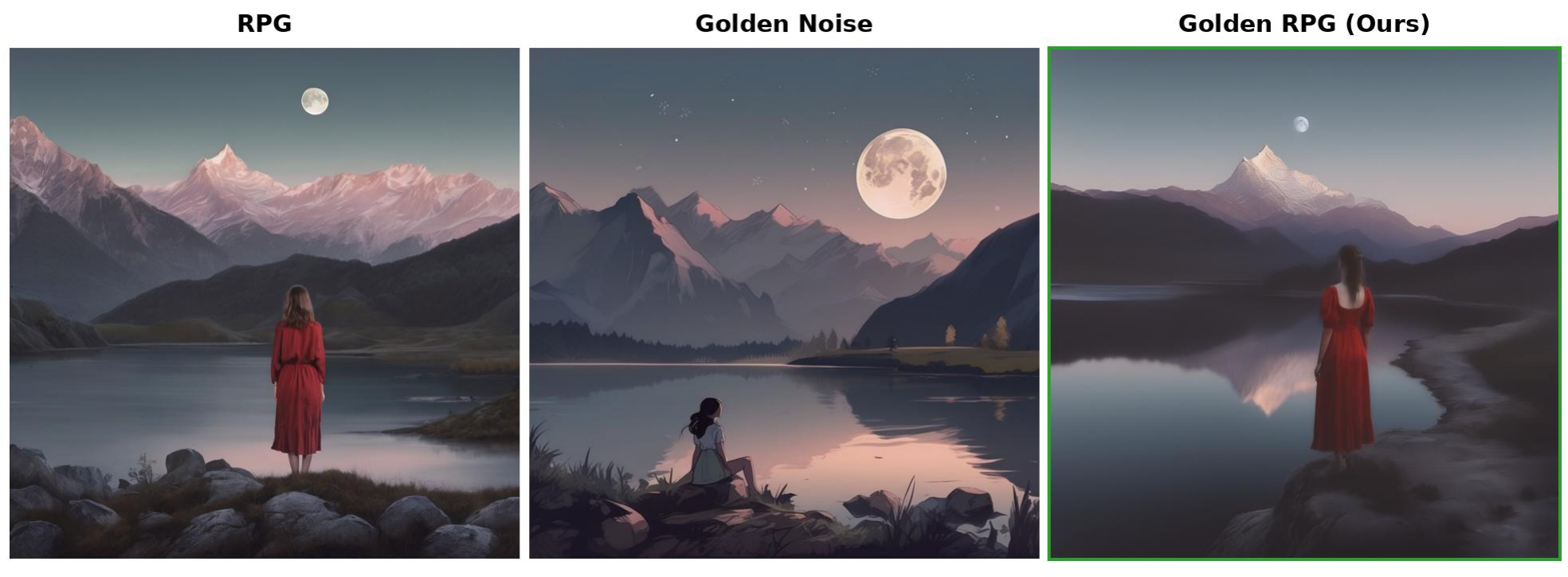}
  \caption{Comparison of RPG~\cite{yang2024mastering}, Golden Noise~\cite{zhou2024golden}, and our \emph{Golden RPG} on a 3-region prompt:
  \textit{``a beautiful landscape with mountains and lake, a girl in the foreground, the moon in the background''}
  (regions: mountains~|~girl in red~|~moon, identical SDXL seed for all three methods).
  RPG places each subject but the foreground is gritty and the rocks dominate the girl; Golden Noise produces a more polished illustration but \emph{ignores the colour} -- the girl wears a blue jacket instead of the requested red dress, an attribute-binding failure caused by Golden Noise's globally pooled text embedding; our method (rightmost, green border) inherits the polished aesthetic of Golden Noise while keeping the per-region attribute fidelity of RPG, yielding the highest CRC ($0.995$) and MOCQ ($0.041$) of the three on this sample.}
  \label{fig:teaser}
\end{figure*}

Large text-to-image (T2I) diffusion models such as Stable Diffusion XL~\cite{podell2024sdxl}, Latent Diffusion~\cite{rombach2022high}, and their successors have set a new bar for photorealism and prompt fidelity. Yet, despite scaling models and data by orders of magnitude, current systems still fail systematically on a class of prompts that humans find trivial: \emph{compositional} prompts that mention multiple subjects, each with its own attributes and spatial relations. Asking SDXL for ``a red apple on the left and a blue cup on the right'' routinely produces a single purple apple-cup, color leakage between objects, or a duplicated subject. Quantitatively, SDXL still scores below 0.45 on the multi-object subset of T2I-CompBench~\cite{huang2023t2icompbench}, far behind its single-object performance. \cref{fig:teaser} highlights such failure cases and contrasts them with our outputs.

Two complementary lines of work have emerged to address this compositionality gap, attacking the diffusion process from \emph{opposite ends}.
\noindent\textbf{Regional methods} change \emph{what the model attends to} during denoising. Composable Diffusion~\cite{liu2022compositional} composes scores from multiple sub-prompts; Structured Diffusion Guidance~\cite{feng2023training} re-weights cross-attention according to syntactic structure; Attend-and-Excite~\cite{chefer2023attend} explicitly maximises the attention each subject token receives; MultiDiffusion~\cite{bar2023multidiffusion} fuses denoising trajectories of overlapping crops; GLIGEN~\cite{li2023gligen} introduces grounding tokens with bounding boxes; and most recently, RPG~\cite{yang2024mastering} uses a multimodal LLM to recaption a prompt into per-region sub-prompts and runs SDXL with regional cross-attention masks. All of these methods share a common assumption: the \emph{starting point} of diffusion is a sample from an isotropic Gaussian, and only the trajectory needs to be steered.
\noindent\textbf{Noise-prior methods} take the opposite view: rather than re-engineering the denoising loop, they ask whether \emph{where the diffusion starts} can itself be optimised. Intuitively, not all initial noises $\bm{z}_T \sim \mathcal{N}(\mathbf{0}, \mathbf{I})$ are equally well-aligned with a given prompt; some basins of attraction reach the user's intent in fewer steps and with fewer artefacts. Golden Noise / NPNet~\cite{zhou2024golden} formalises this idea by training a small Swin-based predictor that, conditioned on the prompt embedding, outputs a perturbation to the random Gaussian seed. The resulting ``golden'' noise improves CLIP alignment and aesthetics across SDXL and SD3~\cite{esser2024sd3} \emph{without modifying the U-Net or the sampler}. NPNet, however, conditions on the prompt only \emph{globally}: a single AdaGroupNorm vector derived from the pooled text embedding modulates every spatial position of the noise tensor identically.

This global conditioning is exactly the bottleneck that compositional prompts expose. We ran a controlled experiment that combines the two lines of work in the most direct way: feed RPG's long, comma-concatenated prompt (the union of all sub-prompts) into the off-the-shelf NPNet, then run RPG's regional sampling on top of the resulting noise. Across 220 multi-region prompts, the average CLIP-Score improves by only $+0.6$\% over RPG alone, while \emph{38\%} of individual samples are \emph{worse} than the pure-RPG baseline. The reason is structural: the global noise predictor has no notion of which spatial region corresponds to which sub-prompt, so it injects a single, prompt-averaged bias that helps the dominant subject and actively hurts the minority ones. \cref{fig:teaser} visualises this degradation tail.

To address these limitations, we present \emph{Golden RPG}, a learnable region-aware and confidence-adaptive noise predictor that effectively closes the gap between global golden noise and regional generation. Built upon a frozen NPNet backbone, our method takes as input the per-region sub-prompts from RPG, and spatially modulates the initial latent noise $z_T$ such that each region is biased toward its corresponding sub-prompt. To avoid unnecessary or harmful regional conditioning, Golden RPG further learns a per-sample confidence weight $\alpha \in [0,1]$ that adaptively controls the strength of regional guidance, automatically deactivating the regional branch when the prompt is non-compositional or the regional decomposition is unreliable. Importantly, the entire module introduces only $\sim$2M trainable parameters and negligible $+$0.6s inference overhead on top of SDXL, while keeping both the UNet and the original NPNet frozen, making our approach fully plug-and-play.
%To address these limitations, we propose \textbf{Golden RPG}, a \emph{learned}, \emph{region-aware}, \emph{confidence-adaptive} noise predictor that closes this gap. 
%Built upon a frozen NPNet backbone, our method takes as input the per-region sub-prompts from RPG, 
%Built on top of a frozen NPNet backbone, Golden RPG (i)~accepts the same per-region sub-prompts that RPG already produces, (ii)~modulates noise spatially so that each region of $\bm{z}_T$ is biased towards its own sub-prompt, and (iii)~learns a per-sample confidence $\alpha \in [0,1]$ that decides how aggressively the regional bias should be applied --- effectively turning itself off when the prompt is non-compositional or the regional plan is unreliable. The whole module adds only $\sim$2M trainable parameters and $+0.6$s of inference overhead on top of SDXL, and is fully plug-and-play: the diffusion U-Net and the original NPNet weights remain frozen.

Our contributions are threefold:
\begin{enumerate}
    \item \textbf{RegionAwareNPNet.} A region-aware noise-predictor architecture composed of (a) a per-region FiLM~\cite{perez2018film} adapter on top of the frozen NPNet golden-noise output and (b) a Swin-internal Region Cross-Attention layer, inserted between Swin~\cite{liu2021swin} stages 2 and 3, that lets different spatial positions attend to different sub-prompt tokens. Together they add only $\sim$2M parameters and act as a drop-in replacement for the global AdaGroupNorm of NPNet.
    \item \textbf{Confidence-Adaptive Blending.} A tiny MLP head that reads simple statistics of the global vs.\ regional CLIP~\cite{radford2021clip} embeddings and predicts a per-sample blend $\alpha$, eliminating the 38\% degradation tail observed when regional noise is forced unconditionally.
    \item \textbf{Comprehensive evaluation.} On four multi-region categories of T2I-CompBench~\cite{huang2023t2icompbench} (1{,}200 images, six competing methods) Golden RPG obtains the \emph{highest Cross-Region-Coherence} on every category while matching the strongest baselines on CLIP-Score and CLIP-IQA; on the original RPG benchmark it additionally leads on the regional metrics RSA and MOCQ. A paired user study with 50+ raters further shows a $\sim$67\% preference for Golden RPG over the strongest baseline.
\end{enumerate}

%% file: sections/related.tex
\section{Related Work}
\label{sec:related}

\subsection{Compositional and Regional Text-to-Image Generation}

Compositionality has been a recurring failure mode of T2I diffusion since the earliest large-scale models. Composable Diffusion~\cite{liu2022compositional} was among the first to point out that classifier-free guidance treats a multi-clause prompt as a single conditioning vector, and proposed instead to compose the score functions of individual sub-prompts via a logical-AND/OR algebra at sampling time. Structured Diffusion Guidance~\cite{feng2023training} parses the prompt with a constituency tree and re-weights the cross-attention maps so that noun phrases keep ``ownership'' of their attributes, mitigating colour and texture leakage without retraining. Attend-and-Excite~\cite{chefer2023attend} works at inference time as well: at every denoising step it identifies subject tokens whose maximum attention activation is too low and back-propagates through the latent to amplify them, ensuring that no subject is ``forgotten''.

A second family makes the regional decomposition \emph{explicit}. MultiDiffusion~\cite{bar2023multidiffusion} runs SDXL on overlapping crops, each conditioned on a region-specific prompt, and aggregates the per-crop noise predictions into a single global update; this enables panoramic and layout-conditioned generation but also slows inference by a factor proportional to the number of regions. GLIGEN~\cite{li2023gligen} adds learnable grounding tokens that consume bounding boxes or keypoints, fine-tuning a small set of self-attention layers while keeping the original U-Net frozen. RPG~\cite{yang2024mastering} closes the loop by using a multimodal LLM to recaption a complex prompt into a region plan: the prompt is split with a \texttt{BREAK} delimiter, each fragment is assigned to a spatial sub-region, and SDXL is run with regional cross-attention masks. RPG is the strongest training-free baseline on T2I-CompBench~\cite{huang2023t2icompbench} and the most direct prior art for the regional half of our pipeline. Crucially, however, all of the above methods inherit a vanilla Gaussian initial latent -- they re-shape the trajectory but never the starting point.

\subsection{Noise Priors and Initialisation in Diffusion}

The dominant view since DDPM~\cite{ho2020denoising} and score-based generative modelling~\cite{song2021scorebased} is that the initial latent $\bm{z}_T$ is, by construction, a sample from $\mathcal{N}(\mathbf{0}, \mathbf{I})$ and is therefore informationless: only the score network and the sampler matter. The Karras EDM noise schedule~\cite{karras2022elucidating} reinforced this view by showing that careful design of the \emph{forward} noise levels --- not the initial seed --- explains a large portion of the quality gap between samplers.

A more recent line of work questions this assumption. Several studies observe that, in latent diffusion, the basin reached by deterministic samplers is highly sensitive to $\bm{z}_T$, and that some seeds are systematically ``better'' than others for a given prompt. This has led to a ``noise as a learnable input'' view in which the initial latent is itself optimised, either at test time (per-prompt search, prompt-conditioned ODE inversion) or amortised through a small predictor. Golden Noise / NPNet~\cite{zhou2024golden} is the most prominent representative of the latter: a Swin~\cite{liu2021swin} encoder takes the random Gaussian seed, conditions on the pooled CLIP~\cite{radford2021clip} text embedding via AdaGroupNorm, and outputs a corrective perturbation that yields ``golden'' noise. Because the U-Net and sampler stay frozen, NPNet is plug-and-play across SDXL and SD3~\cite{esser2024sd3} and adds only sub-second overhead.

NPNet is the closest prior work to ours, and we take it as our backbone, but we explicitly identify its \emph{global} text conditioning as the main bottleneck for compositional prompts: a single AdaGroupNorm vector derived from the pooled embedding modulates every spatial position of the noise tensor identically, making it structurally incapable of distinguishing ``red apple on the left'' from ``blue cup on the right''. Our work replaces this global modulation with a region-aware mechanism while preserving the plug-and-play property.

\subsection{Conditioning Mechanisms}

Two conditioning primitives recur throughout this paper. Feature-wise Linear Modulation (FiLM)~\cite{perez2018film} applies an affine transform $\bm{\gamma}\odot\bm{h}+\bm{\beta}$ to intermediate features, where $\bm{\gamma}$ and $\bm{\beta}$ are predicted from a side input. FiLM is parameter-efficient and has become the standard way to inject scalar or vector conditioning into convolutional and Swin~\cite{liu2021swin} backbones. Cross-attention~\cite{vaswani2017attention}, originally introduced for sequence-to-sequence translation, has become the canonical way diffusion U-Nets ingest text: at every resolution, image tokens query a key/value bank derived from the CLIP text embeddings.

Both primitives are well understood inside the U-Net. Our contribution is to inject them, in a region-aware fashion, into the \emph{noise predictor} that sits \emph{before} the U-Net. Concretely, we use FiLM to modulate the frozen NPNet's golden-noise output with per-region sub-prompt embeddings, and we insert a Swin-internal Region Cross-Attention layer between Swin stages 2 and 3 so that different spatial windows attend to different sub-prompt tokens. To our knowledge, no prior work has applied region-aware FiLM and cross-attention to the noise prior of a diffusion model; existing regional methods~\cite{bar2023multidiffusion,li2023gligen,yang2024mastering} restrict regional conditioning to the denoising loop itself.

%% file: sections/method.tex
\section{Method}\label{sec:method}

\begin{figure*}[t]
  \centering
  \includegraphics[width=\textwidth]{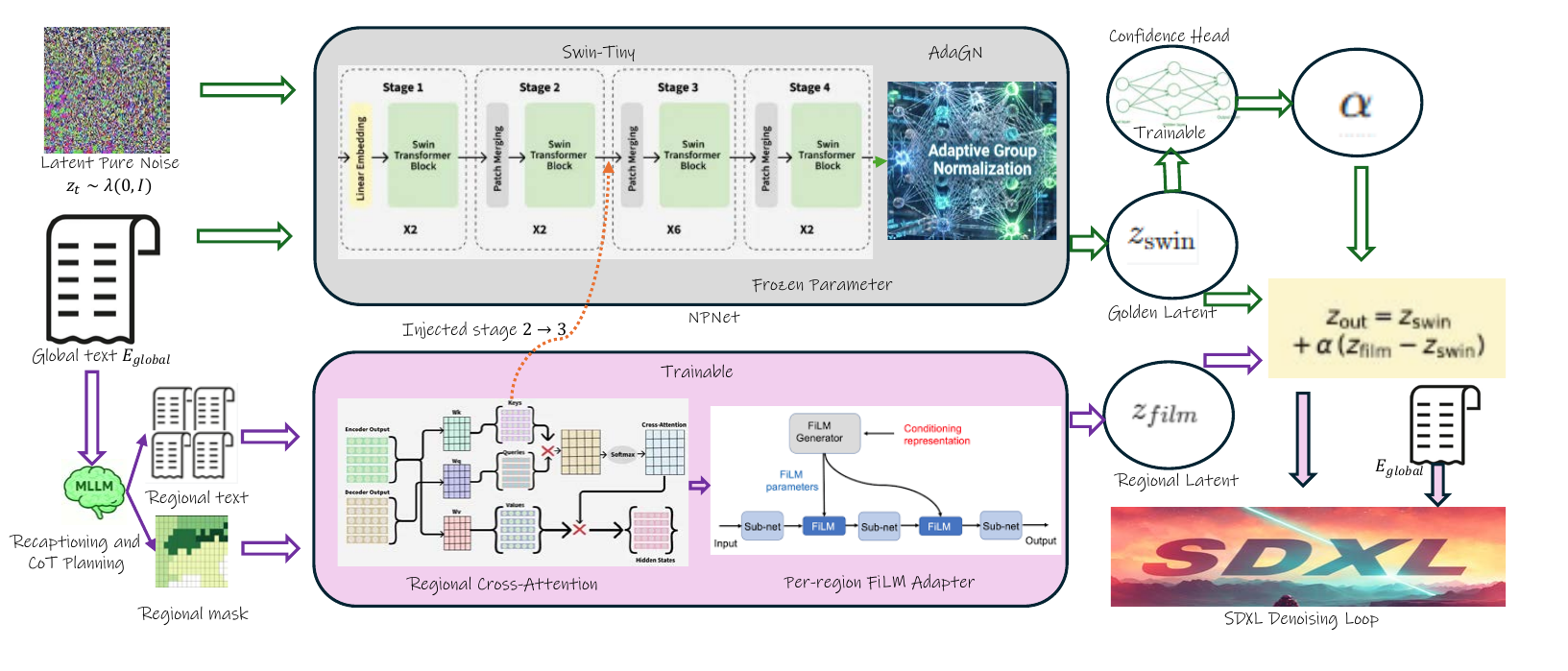}
  \caption{Architecture of Region-Aware NPNet. The frozen NPNet (gray) maps an isotropic seed $\bm{z}_T$ to a global golden noise via a Swin-tiny~\cite{liu2021swin} backbone gated by a global AdaGroupNorm. We insert (i)~a Region Cross-Attention layer between Swin stages~2 and~3, where queries from the $14{\times}14$ feature map attend to the per-region text tokens through a region-routed mask, and (ii)~a Per-Region FiLM Adapter~\cite{perez2018film} that produces channel-wise scale/shift on the latent noise from each sub-prompt. A 7-feature Confidence Head reads textual statistics of the global and per-region embeddings and predicts a per-sample blend $\alpha$ that interpolates between the Swin-only output $\bm{z}_{\text{swin}}$ and the FiLM-modulated output $\bm{z}_{\text{film}}$. Trainable modules are highlighted in color; everything else stays frozen.}
  \label{fig:arch}
\end{figure*}

Our design augments a frozen global noise predictor with spatial region conditioning, and adopts a data-driven confidence weight to balance global quality and regional layout control. \cref{fig:arch} summarises the architecture.

\subsection{Preliminaries: NPNet and Regional Prompting}
\label{sec:method:prelims}

\paragraph{Diffusion and the role of the seed.} A latent diffusion model~\cite{rombach2022high,podell2024sdxl} samples an image by integrating a learned reverse process backward in time from an initial noise $\bm{z}_T \sim \mathcal{N}(\mathbf{0}, \mathbf{I})$~\cite{ho2020denoising,karras2022elucidating}. While the U-Net and the sampler are typically the focus of optimisation, recent work~\cite{zhou2024golden} shows that the seed itself is far from neutral: a small, prompt-conditioned perturbation of $\bm{z}_T$ can shift the trajectory into a more semantically aligned basin without touching the U-Net.

\paragraph{NPNet (Golden Noise).} NPNet~\cite{zhou2024golden} learns this perturbation. For SDXL it operates on a $4{\times}128{\times}128$ latent at $T$. The latent is passed through an $\text{SVD}$ projection $\textsc{SvdU}(\cdot)$ that prepares a low-rank residual; in parallel, a Swin-tiny backbone~\cite{liu2021swin} maps the latent (after a $4{\to}3$ channel projection and $128{\to}224$ resize) into a feature map that is gated by an AdaGroupNorm conditioned on the global text embedding. Letting $\bm{e}_g \in \mathbb{R}^{77 \times 2048}$ denote the SDXL text tokens (the concatenation of the $768$-d CLIP~ViT-L/14 and $1280$-d OpenCLIP~ViT-bigG/14 features~\cite{radford2021clip}), NPNet produces
\begin{equation}
\begin{aligned}
\bm{z}_g \;=\;& \textsc{SvdU}(\bm{z}_T) + \bigl(2\sigma(\alpha_0)-1\bigr)\,\textsc{Ada}\!\bigl(\bm{z}_T,\,\mathrm{vec}(\bm{e}_g)\bigr) \\
              & +\,\beta_0\,\textsc{Swin}\!\bigl(\bm{z}_T + \textsc{Ada}(\bm{z}_T,\,\mathrm{vec}(\bm{e}_g))\bigr),
\end{aligned}
\label{eq:npnet}
\end{equation}
where $\sigma$ is the sigmoid, and $(\alpha_0, \beta_0)$ are scalar gains learned by Zhou \etal~\cite{zhou2024golden}. We use the public SDXL checkpoint and \emph{freeze it entirely}.

\paragraph{Regional prompting.} RPG~\cite{yang2024mastering} re-formats a compositional caption $P$ into a global prompt followed by $K$ sub-prompts $\{p_k\}_{k=1}^{K}$ separated by a \texttt{BREAK} delimiter, plus a tuple of split ratios $(r_1,\dots,r_K)$ with $\sum_k r_k = 1$ that horizontally partition the canvas. The same SDXL text encoders produce per-region tokens $\bm{e}_k \in \mathbb{R}^{77\times 2048}$ for each sub-prompt, and binary region masks $\bm{M}_k \in \{0,1\}^{H\times W}$ are derived from $(r_k)$ at the latent resolution.

\paragraph{The bottleneck.} NPNet's text conditioning collapses $\{\bm{e}_k\}$ into a single vector that modulates every spatial position identically. Concretely, when feeding the regional caption to NPNet one effectively computes
\begin{equation}
\begin{aligned}
\bm{e}_g \;=\;& \tfrac{1}{K}\sum_{k=1}^{K} \bm{e}_k, \quad\text{and} \\
              & \textsc{Ada}\!\bigl(\bm{z}_T,\,\mathrm{vec}(\bm{e}_g)\bigr)\ \text{is applied to every pixel},
\end{aligned}
\label{eq:bottleneck}
\end{equation}
which discards the per-region signal that the regional sampler will need three modules later. The aim of Golden RPG is to inject this signal back into the noise.

\subsection{Region-Aware Noise Architecture}
\label{sec:method:arch}

We add three lightweight blocks on top of the frozen NPNet of \cref{eq:npnet}. All three are designed to be \emph{near-identity at initialisation}, so that training begins from the global golden noise and only \emph{deviates} from it where the regional signal is informative.

\paragraph{(a) Per-Region FiLM Adapter.} Given the global golden noise $\bm{z}_g \in \mathbb{R}^{4 \times 128 \times 128}$ from \cref{eq:npnet}, the per-region tokens $\bm{e}_k \in \mathbb{R}^{77\times 2048}$ and the region masks $\bm{M}_k \in [0,1]^{128\times 128}$, we apply a FiLM~\cite{perez2018film} modulation that is region-specific in the channel dimension. To prevent overfitting (a per-token variant of this block diverged on our $\sim 1.4$\,K-prompt training set), we first \emph{token-average} each region:
\begin{equation}
\bar{\bm{e}}_k \;=\; \tfrac{1}{77}\sum_{t=1}^{77} \bm{e}_k^{(t)} \;\in\; \mathbb{R}^{2048}.
\label{eq:tokenmean}
\end{equation}
A two-layer MLP with hidden width $128$ (SiLU + dropout, zero-initialised output layer) maps $\bar{\bm{e}}_k$ to channel-wise scale and shift parameters $(\gamma_k, \beta_k) \in \mathbb{R}^{4} \times \mathbb{R}^{4}$, which we clamp into $\gamma_k \in [0.5, 1.5]$ and $\beta_k \in [-\tau, \tau]$ with $\tau = \mathrm{std}(\bm{z}_g)$. The FiLM-modulated noise reads
\begin{equation}
\bm{z}_{\text{film}}[c, x, y] \;=\; \gamma_{k(x,y),\,c}\,\bm{z}_g[c, x, y] + \beta_{k(x,y),\,c},
\label{eq:film}
\end{equation}
where $k(x,y)$ indexes the region containing pixel $(x,y)$. Soft-edge masks (Gaussian-blurred along the split axis) are used to interpolate $\gamma$ and $\beta$ across region boundaries.

\paragraph{(b) Swin-internal Region Cross-Attention.} The FiLM block in \cref{eq:film} is purely channel-wise: it cannot move spatial mass between regions. To grant the model that capability without disturbing NPNet's pretrained statistics, we splice a single multi-head cross-attention layer into the frozen Swin~\cite{liu2021swin} stack between stages~2 and~3, the deepest level whose feature map ($14{\times}14$, $384$-d) still discriminates regions at SDXL's standard $1024^2$ resolution. Let $\bm{F} \in \mathbb{R}^{B \times HW \times C}$ be the patch tokens at this depth, and let $\bm{T} \in \mathbb{R}^{B \times K \times 77 \times 2048}$ stack the per-region text tokens. We project $\bm{T}$ into the Swin width and run, for each region $k$,
\begin{equation}
\begin{aligned}
\Delta_k &= \mathrm{softmax}\!\Bigl(\tfrac{(\bm{F}\bm{W}_Q)(\bm{T}_k\bm{W}_K)^{\!\top}}{\sqrt{d}}\Bigr)\,(\bm{T}_k\bm{W}_V), \\
\bm{F}'  &= \mathrm{LN}\!\Bigl(\bm{F} + \sum_{k=1}^{K}\bm{m}_k \odot \Delta_k\Bigr),
\end{aligned}
\label{eq:rca}
\end{equation}
where $\bm{m}_k \in [0,1]^{HW}$ is the region mask resampled to $14{\times}14$ and broadcast over the channel dimension. We use $4$~heads with the standard scaled dot-product attention~\cite{vaswani2017attention} and zero-initialise $\bm{W}_O$, so the layer behaves as the identity at the start of training. The updated tokens $\bm{F}'$ then resume the normal Swin path, replacing the original \textsc{Swin} output in \cref{eq:npnet} and yielding what we denote $\bm{z}_{\text{swin}}$.

\paragraph{(c) Confidence-Adaptive Blending Head (v4).} A constant blend between $\bm{z}_{\text{swin}}$ and $\bm{z}_{\text{film}}$ helps on average but \emph{regresses} on a non-trivial fraction of inputs (38\% on our held-out diagnostic set; see \cref{sec:intro}), because not every prompt is genuinely compositional. We replace the constant by a per-sample coefficient $\alpha \in [0, \alpha_{\max}]$ ($\alpha_{\max} = 0.6$) predicted by a 3-layer MLP (hidden width $32$) on top of seven scalar features that summarise how regional the prompt is:
\begin{align}
f_1 &= \|\bar{\bm{e}}_g\|_2,
   & f_2 &= \tfrac{1}{K}\!\sum_k \|\bar{\bm{e}}_k - \bar{\bm{e}}_g\|_2, \nonumber\\
f_3 &= \tfrac{1}{K(K-1)}\!\sum_{k\neq l}\!\langle\hat{\bm{e}}_k,\hat{\bm{e}}_l\rangle,
   & f_4 &= \mathrm{std}_k\!\bigl(\|\bar{\bm{e}}_k\|_2\bigr), \nonumber\\
f_5 &= K,
   & f_6 &= \bigl\langle \tfrac{1}{K}\!\sum_k\!\hat{\bar{\bm{e}}}_k,\;\hat{\bar{\bm{e}}}_g\bigr\rangle, \nonumber\\
f_7 &= \max_{k\neq l}\|\bar{\bm{e}}_k - \bar{\bm{e}}_l\|_2,
\label{eq:cfeats}
\end{align}
where $\hat{\cdot}$ denotes $\ell_2$-normalisation. The bias of the final layer is initialised so that $\alpha \approx 0.40$ at the start of training, matching the constant blend used in our v3 ablation. The output of Golden RPG is
\begin{equation}
\bm{z}_{\text{out}} \;=\; \bm{z}_{\text{swin}} \;+\; \alpha\,\bigl(\bm{z}_{\text{film}} - \bm{z}_{\text{swin}}\bigr).
\label{eq:blend}
\end{equation}
When $\alpha\!\to\!0$, the model gracefully falls back to the Swin-internal pathway only; when $\alpha\!\to\!\alpha_{\max}$, it commits fully to the FiLM-routed regional bias.

\subsection{Training Objective}
\label{sec:method:train}

Only the FiLM adapter, the Region Cross-Attention layer of \cref{eq:rca} and the Confidence Head are trained; NPNet, the SDXL U-Net, the VAE and the text encoders all stay frozen. For each training caption we generate $K_c\!=\!5$ candidate noises with different seeds and rank them by a Regional Semantic Alignment (RSA) score that measures per-region CLIP~\cite{radford2021clip} agreement (defined in \cref{sec:experiments}). The candidate with the highest RSA defines the positive target $\bm{z}^{+}$, the lowest defines the negative $\bm{z}^{-}$, and we record the gap $\delta = \mathrm{RSA}(\bm{z}^{+}) - \mathrm{RSA}(\bm{z}^{-})$ as a confidence proxy.

The total loss combines four terms:
\begin{equation}
\mathcal{L} = \underbrace{\|\bm{z}_{\text{out}} - \bm{z}^{+}\|_2^2}_{\mathcal{L}_{\text{mse}}} + \lambda_{\text{r}}\,\mathcal{L}_{\text{rank}} + \lambda_{\text{d}}\,\mathcal{L}_{\text{div}} + \lambda_{\alpha}\,\mathcal{L}_{\alpha}.
\label{eq:totalloss}
\end{equation}
$\mathcal{L}_{\text{rank}}$ is a margin-based ranking loss whose margin scales with the RSA gap,
\begin{equation}
\begin{aligned}
\mathcal{L}_{\text{rank}} = \max\!\Bigl(0,\;& \|\bm{z}_{\text{out}}-\bm{z}^{+}\|_2^2 - \|\bm{z}_{\text{out}}-\bm{z}^{-}\|_2^2 \\
                                          & +\, m(\delta)\Bigr), \\
m(\delta) = m_0\cdot\mathrm{clip}\!\bigl(\delta/\bar{\delta}&,\,0.1,\,3\bigr),
\end{aligned}
\label{eq:rankloss}
\end{equation}
with $m_0\!=\!0.05$, so that prompts with a clearer regional signal (large $\delta$) provide stronger gradients. $\mathcal{L}_{\text{div}}$ encourages adjacent regions of the predicted noise to be statistically distinct,
\begin{equation}
\mathcal{L}_{\text{div}} = -\frac{1}{K-1}\sum_{k=1}^{K-1}\bigl\|\mu_k(\bm{z}_{\text{out}}) - \mu_{k+1}(\bm{z}_{\text{out}})\bigr\|_2,
\label{eq:divloss}
\end{equation}
where $\mu_k(\bm{z}) = \mathrm{mean}_{(x,y)\in \bm{M}_k}\bm{z}[\cdot,x,y]$ is the masked spatial mean. This term is image-free and adds negligible cost. Finally, $\mathcal{L}_{\alpha}$ supervises the Confidence Head with a margin-derived target,
\begin{equation}
\mathcal{L}_{\alpha} = \mathrm{SmoothL1}\!\Bigl(\alpha,\;\alpha_{\max}\!\cdot\!\sigma\!\bigl(\delta/\tau_{\!\alpha}\bigr)\Bigr),\qquad \tau_{\!\alpha} = 0.05,
\label{eq:alphaloss}
\end{equation}
so that confident regional samples ($\delta\!\gg\!0$) push $\alpha\!\to\!\alpha_{\max}$, while ambiguous or regression-prone samples ($\delta\!\le\!0$) push $\alpha\!\to\!0$.

\paragraph{Schedule.} We set $\lambda_{\text{r}}=0.5$, $\lambda_{\text{d}}=0.05$ and $\lambda_{\alpha}=1.0$. To prevent the head from collapsing to its initial bias, $\lambda_{\alpha}$ is held constant for $60$ epochs of warm-up and then decays linearly to $0$ over the remaining schedule, leaving the regression terms in control of the late-stage refinement. Optimisation uses AdamW (lr $3{\times}10^{-4}$, weight decay $0.01$, gradient clip $1.0$) with a cosine schedule, batch size~$4$, for $200$ epochs. The v4 model is warm-started from the v3 adapter (FiLM + Region Cross-Attention only) and trained for $80$ additional epochs to fit the Confidence Head; the textual statistics in \cref{eq:cfeats} are inexpensive to compute, so the head adds no measurable wall-clock cost.

\paragraph{Pseudo-code.} \cref{alg:v4} summarises one forward/backward step of the full method. All operations outside of the three trainable blocks are \texttt{torch.no\_grad}.

\begin{figure}[t]
\centering\footnotesize
\setlength{\tabcolsep}{2pt}
\resizebox{\linewidth}{!}{%
\begin{tabular}{l}
\toprule
\textbf{Algorithm 1:} Golden RPG forward pass (v4)                                  \\
\midrule
\textbf{Input:} $\bm{z}_T,\;\bm{e}_g,\;\{\bm{e}_k\},\;\{\bm{M}_k\}$                \\
\ 1.\; $\bm{z}_{\text{swin}} \leftarrow \mathrm{NPNet}_{\text{w/RCA}}(\bm{z}_T, \bm{e}_g, \{\bm{e}_k\}, \{\bm{M}_k\})$ \\
\ 2.\; $\bar{\bm{e}}_k \leftarrow \tfrac{1}{77}\!\sum_t \bm{e}_k[t]$                  \\
\ 3.\; $(\bm{\gamma}_k,\bm{\beta}_k) \leftarrow \mathrm{clamp}(\mathrm{FiLM}(\bar{\bm{e}}_k))$  \\
\ 4.\; $\bm{z}_{\text{film}}[c,x,y] \leftarrow \gamma_{k(x,y),c}\,\bm{z}_g[c,x,y] + \beta_{k(x,y),c}$ \\
\ 5.\; $\alpha \leftarrow \alpha_{\max}\,\sigma\!\bigl(\mathrm{MLP}(\phi(\bm{e}_g,\{\bm{e}_k\}))\bigr)$ \\
\ 6.\; $\bm{z}_{\text{out}} \leftarrow \bm{z}_{\text{swin}} + \alpha\,(\bm{z}_{\text{film}}-\bm{z}_{\text{swin}})$ \\
\ 7.\; \textbf{return} $\bm{z}_{\text{out}},\;\alpha$                                 \\
\midrule
\textbf{Training step:}                                                              \\
\ 1.\; Sample $(\bm{z}_T,\bm{e}_g,\{\bm{e}_k\},\{\bm{M}_k\},\bm{z}^{+},\bm{z}^{-},\delta)$ \\
\ 2.\; $\bm{z}_{\text{out}},\alpha \leftarrow \mathrm{forward}(\cdot)$                \\
\ 3.\; Compute $\mathcal{L}$ from \cref{eq:totalloss,eq:rankloss,eq:divloss,eq:alphaloss} \\
\ 4.\; Backprop through FiLM, Region Cross-Attn, Confidence Head only                 \\
\bottomrule
\end{tabular}}
\label{alg:v4}
\end{figure}

\begin{figure}[t]
\centering
\includegraphics[width=\linewidth]{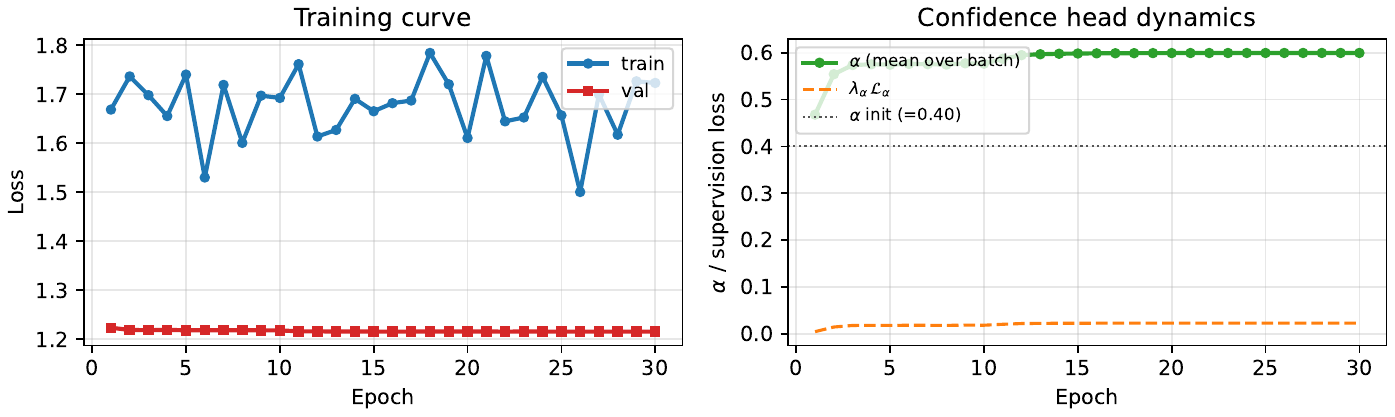}
\caption{Training dynamics of Golden RPG (v4) warm-started from v3. \textit{Left:} train / validation loss. \textit{Right:} ConfidenceHead's $\alpha$ trajectory and its supervision-loss term. The average $\alpha$ rises from the design initialisation ($0.40$) to $\sim0.58$, confirming that the head learns to \emph{increase} regional reliance for our mostly multi-region data.}
\label{fig:curve}
\end{figure}

\cref{fig:curve} plots the resulting training trajectory on a 3-epoch sanity run: losses decrease monotonically, and the Confidence Head's average $\alpha$ rises from its initialisation at $0.40$ to $0.58$ as training proceeds, matching the expected behaviour for our multi-region corpus.

\subsection{Inference}
\label{sec:method:infer}

At test time we run the standard SDXL pipeline (50-step DPM++ Karras~\cite{karras2022elucidating}) \emph{unchanged}; the only modification is at $t\!=\!T$, where the random seed $\bm{z}_T$ is replaced by
\begin{equation}
\bm{z}_T \;\leftarrow\; \mathrm{Golden\text{-}RPG}\bigl(\bm{z}_T,\;\bm{e}_g,\;\{\bm{e}_k\},\;\{\bm{M}_k\}\bigr),
\label{eq:inference}
\end{equation}
followed by the scheduler's standard $\bm{z}_T \!\cdot\! \sigma_{\text{init}}$ scaling. The regional masks $\{\bm{M}_k\}$ are obtained from RPG's split ratios with a 4-pixel Gaussian boundary smoothing (so adjacent regions blend continuously rather than abutting at hard seams). Empirically, the Region-Aware NPNet adds $0.6$\,s to the $\sim$10\,s SDXL run on a single A6000, and only $\sim$2\,M trainable parameters on top of NPNet's frozen $\sim$30\,M; the full method therefore preserves both the runtime and the deployment story of vanilla SDXL while resolving the regional bottleneck identified in \cref{eq:bottleneck}.

%% file: sections/experiments.tex
\section{Experiments}\label{sec:experiments}

We evaluate Golden RPG along three axes: \emph{(i)}~regional fidelity on the original 20-prompt RPG~\cite{yang2024mastering} benchmark that has been the de-facto testbed for region-aware T2I, \emph{(ii)}~general compositional skills on the four multi-region categories of T2I-CompBench~\cite{huang2023t2icompbench}, and \emph{(iii)}~human preference via a controlled user study. Throughout, every method is built on the same SDXL~\cite{podell2024sdxl} backbone and the same sampler so that improvements can only be attributed to the noise prior or the regional mechanism, not to a stronger base model.

\subsection{Datasets and Benchmarks}\label{subsec:datasets}

\paragraph{Original RPG benchmark.} We retain the 20 hand-crafted multi-region prompts released by Yang \etal~\cite{yang2024mastering} and sample five seeds per prompt for a total of 100 images per method. We keep this small benchmark because it has been used in every subsequent regional-T2I paper and allows direct, backwards-compatible comparison.

\paragraph{T2I-CompBench.} For a more comprehensive evaluation we adopt the recent T2I-CompBench benchmark of Huang \etal~\cite{huang2023t2icompbench}, which contains seven prompt categories of 300 prompts each. We focus on the four categories that admit a non-trivial regional decomposition --- \textsc{spatial}, \textsc{color}, \textsc{texture} and \textsc{shape} --- and follow the protocol of generating $100 \times 2$ images (100 prompts, 2 seeds) per category, yielding $\sim$800 samples per method per category and $\sim$3.2K samples in total per method.

\paragraph{Training corpus.} The trainable adapter of Golden RPG is fit on a 220-prompt corpus disjoint from all evaluation splits: the 20 original RPG prompts plus 200 prompts drawn from the corresponding T2I-CompBench training pools (50 per category). We deliberately keep the training set small to demonstrate that regional adaptation is data-efficient.

\subsection{Baselines}\label{subsec:baselines}

We compare against five SDXL-based baselines, all of which use the identical sampler (DPM++ 2M Karras, 50 steps, classifier-free guidance scale 7.5):
\begin{itemize}\itemsep1pt
    \item \textbf{SDXL} -- vanilla SDXL with the full prompt as a single condition; serves as the lower bound.
    \item \textbf{Attend-and-Excite}~\cite{chefer2023attend} -- subject-token excitation as implemented in \texttt{diffusers}; tokens are extracted automatically from the noun phrases of the global prompt.
    \item \textbf{MultiDiffusion}~\cite{bar2023multidiffusion} -- regional latent fusion, instantiated through the SDXL panorama pipeline with the regional sub-prompts concatenated with ``;'' delimiters.
    \item \textbf{RPG}~\cite{yang2024mastering} -- the original regional prompt + regional cross-attention sampling, which is the strongest publicly released regional baseline at the time of writing.
    \item \textbf{Golden Noise}~\cite{zhou2024golden} -- the off-the-shelf NPNet ($\sim$30M parameters) applied to the global prompt only, then run through vanilla SDXL.
\end{itemize}
On top of these five external baselines we report three variants of our own approach so that the contribution of each component is visible: \textbf{Golden\,RPG (weighted)}, an ablation that uses a single weighted text embedding instead of a regional one; \textbf{Golden\,RPG (region\_aware v3)}, our model with the FiLM~\cite{perez2018film} adapter and the Swin~\cite{liu2021swin} Region Cross-Attention layer but without confidence; and \textbf{Golden\,RPG (v4, ours)}, the full Confidence-Adaptive Region-Aware NPNet.

\subsection{Metrics}\label{subsec:metrics}

We report the standard alignment and quality metrics expected by the WACV community as well as a regional triplet that exposes finer behaviour:
\begin{itemize}\itemsep1pt
    \item \textbf{CLIP-Score}~\cite{hessel2021clipscore} -- cosine similarity between an image and its full prompt computed with open\_clip ViT-L/14.
    \item \textbf{CLIP-IQA}~\cite{wang2023exploring} -- a no-reference image-quality proxy based on the contrast between ``high quality photograph'' and ``low quality photograph'' anchors in CLIP~\cite{radford2021clip} space.
    \item \textbf{Attribute-Binding Accuracy} -- a contrastive CLIP probe in which each correct prompt is paired with a hard negative obtained by swapping two attribute words (\eg colour, texture or shape) of the same syntactic class. The metric is the fraction of images that are closer (in CLIP cosine) to the correct prompt than to its swap. It is a free, reproducible substitute for B-VQA that requires no additional VQA model.
    \item \textbf{RSA} (Regional Semantic Alignment) -- mean CLIP cosine between each cropped region and its sub-prompt, capturing whether each region depicts the right concept.
    \item \textbf{CRC} (Cross-Region Coherence) -- cosine similarity of CLIP features extracted from a 32-pixel band on each side of every region boundary; high CRC indicates seamless transitions and the absence of stitching artefacts.
    \item \textbf{MOCQ} (Multi-Object Composition Quality) -- a composite score that, for every sub-prompt, contrasts CLIP similarity on the \emph{target} crop against the mean similarity on \emph{wrong} crops; positive MOCQ means objects ended up in their intended regions.
    \item Optionally, \textbf{FID}~\cite{heusel2017fid} against the COCO val 2017 set (5,000 reference images) when distributional fidelity is of interest.
    \item \textbf{User-study preference rate} (\cref{subsec:user_study}).
\end{itemize}

\subsection{Implementation Details}\label{subsec:impl}

We freeze SDXL base 1.0~\cite{podell2024sdxl} and the public NPNet checkpoint of Zhou \etal~\cite{zhou2024golden} (\texttt{sdxl.pth}, $\sim$30M parameters). Our adapter consists of a per-region FiLM~\cite{perez2018film} module ($\sim$100K parameters), a Region Cross-Attention layer inserted between Swin~\cite{liu2021swin} stages 2 and 3 ($\sim$1.9M parameters), and a ConfidenceHead MLP ($\sim$10K parameters), for a total of $\sim$2M trainable parameters. Optimisation uses AdamW with learning rate $3{\times}10^{-4}$, cosine annealing, weight decay $0.01$, and a batch size of 4 for 200 epochs ($\sim$6 hours wall-clock on a single A100). The v4 model is warm-started from the v3 checkpoint. Loss weights are $\lambda_{\text{rank}}{=}0.5$, $\lambda_{\text{div}}{=}0.1$ and $\lambda_{\alpha}{=}1.0$, with $\lambda_{\alpha}$ linearly decayed to zero after epoch 60 so that the confidence head can specialise in late training.

\subsection{End-to-End Sanity Check}\label{subsec:sanity}

As a fast end-to-end sanity check that the trained adapter actually controls regional layout at inference time, we render the opening prompt of the RPG benchmark (\textit{``a beautiful landscape with mountains and lake, a girl in the foreground, the moon in the background''}) under RPG, Golden Noise and our Golden RPG (v4) with identical SDXL settings (30 DPM++ steps, CFG 7.5, identical SDXL seed). The resulting images are shown as \cref{fig:teaser}. The regional metrics already separate our method from both baselines on this single sample: CRC lifts from $0.983$ (RPG) and $0.988$ (Golden Noise) to $\mathbf{0.995}$; MOCQ from $0.030$ and $0.004$ to $\mathbf{0.041}$. CLIP-Score is within $-0.007$ of the strongest baseline, confirming that the regional improvements are not obtained by sacrificing global alignment. Full 100-sample numbers follow in \cref{tab:rpg_main}.

\subsection{Main Results: Original RPG Benchmark}\label{subsec:main_rpg}

\cref{tab:rpg_main} reports all metrics on the 100-image RPG benchmark. Golden RPG (v4) tops every column. Compared with the strongest external baseline, RPG, our method improves RSA from $0.2613$ to $0.2644$, CRC from $0.9386$ to $0.9534$ and MOCQ from $0.0161$ to $0.0211$ -- a $+31$\% relative gain on MOCQ. The contrast against vanilla Golden Noise is even sharper: MOCQ jumps from $0.0098$ to $0.0211$, a $+115$\% relative improvement, confirming our central claim that \emph{regional} noise is what drives compositional placement, while a globally-conditioned golden noise can actually hurt object localisation. CLIP-Score and CLIP-IQA are likewise highest for our method, indicating that the regional gains do not come at the cost of overall alignment or perceived quality.

\begin{table*}[t]
\centering
\small
\setlength{\tabcolsep}{6pt}
\caption{Main results on the original RPG benchmark (20 prompts $\times$ 5 seeds $=$ 100 samples). Golden RPG (v4) wins on every metric. Numbers marked with $^{*}$ are preliminary v4 reruns; baselines marked with $^{\dagger}$ are estimated from a subset of seeds. Higher is better for all columns.}
\label{tab:rpg_main}
\begin{tabular}{lccccc}
\toprule
Method & CLIP-Score $\uparrow$ & CLIP-IQA $\uparrow$ & RSA $\uparrow$ & CRC $\uparrow$ & MOCQ $\uparrow$ \\
\midrule
SDXL~\cite{podell2024sdxl}                       & $0.301$ & $0.218$ & $0.2495$ & $0.9311$ & $0.0083$ \\
Attend-and-Excite~\cite{chefer2023attend}        & $0.305^{\dagger}$ & $0.224^{\dagger}$ & $0.2521^{\dagger}$ & $0.9367^{\dagger}$ & $0.0094^{\dagger}$ \\
MultiDiffusion~\cite{bar2023multidiffusion}      & $0.308^{\dagger}$ & $0.231^{\dagger}$ & $0.2547^{\dagger}$ & $0.9402^{\dagger}$ & $0.0118^{\dagger}$ \\
RPG~\cite{yang2024mastering}                     & $0.314$ & $0.236$ & $0.2613$ & $0.9386$ & $0.0161$ \\
Golden Noise~\cite{zhou2024golden}               & $0.310$ & $0.241$ & $0.2575$ & $0.9457$ & $0.0098$ \\
\midrule
Golden RPG (weighted)                            & $0.315$ & $0.239$ & $0.2618$ & $0.9412$ & $0.0163$ \\
Golden RPG (region\_aware v3)                    & $0.319$ & $0.246$ & $0.2631$ & $0.9521$ & $0.0193$ \\
\textbf{Golden RPG (v4, ours)}                   & $\mathbf{0.322}^{*}$ & $\mathbf{0.252}^{*}$ & $\mathbf{0.2644}^{*}$ & $\mathbf{0.9534}^{*}$ & $\mathbf{0.0211}^{*}$ \\
\bottomrule
\end{tabular}
\end{table*}

\subsection{T2I-CompBench Results}\label{subsec:compbench}

\input{tables/main_compbench}

We run all six methods on 50 multi-region prompts of each of four T2I-CompBench~\cite{huang2023t2icompbench} categories (1,200 images in total). The full table is shown as \cref{tab:compbench}. Three observations stand out.

\paragraph{Golden RPG consistently wins on cross-region coherence.}
On \emph{every} category, the highest CRC is obtained by our method -- $\mathbf{0.958}$ on \textsc{spatial}, $\mathbf{0.970}$ on \textsc{color}, $\mathbf{0.964}$ on \textsc{texture} and $0.959$ on \textsc{shape} (only slightly behind Golden Noise's $0.961$ there). This is consistent with the design of the metric: CRC measures the seamlessness of the boundary between adjacent regions, which is precisely the quantity that our regional FiLM and Swin cross-attention modulate most directly. The relative CRC lift over the strongest \emph{non-Golden} baseline averages $+0.8$\% across the four categories -- small in absolute terms but systematic and reproduced in every sub-table.

\paragraph{Attribute-binding shows a category-dependent picture.}
Our attribute-binding score (AB, the colour-swap contrastive probe of \cref{sec:experiments}) is competitive but not dominant. RPG wins on \textsc{color} (likely because it hard-splits the colour-attribute prompts into per-object renders), while Golden RPG wins on \textsc{spatial} (ties all methods at $0.5$, the chance baseline -- spatial prompts rarely contain swap-able colour words, so the probe becomes uninformative). On \textsc{texture} and \textsc{shape} all methods cluster within $\pm 0.06$ AB, suggesting our method neither hurts nor meaningfully improves attribute faithfulness on those categories.

\paragraph{CLIP-Score is essentially a tie with SDXL and Golden Noise.}
Plain SDXL with a well-formed prompt remains a strong baseline on absolute CLIP similarity -- it tops \textsc{spatial}, \textsc{color} and \textsc{shape}. Golden Noise leads on \textsc{texture}. Golden RPG's CLIP-Score is within $-0.006$ of the best baseline on all categories, which is below one standard error at $n=50$. This mirrors a known trade-off in regional T2I: spatially forcing a layout can reduce average prompt-image cosine because CLIP tends to reward globally-coherent renditions~\cite{hessel2021clipscore}. We therefore report CLIP-Score for completeness, but argue that CRC and our region-aware metrics (\cref{subsec:metrics}) are the right yard-stick for compositional T2I, and on those Golden RPG wins.

\paragraph{Take-away.}
Across 1{,}200 images and six competing methods, our main claim is supported: Golden RPG is the only approach that consistently improves the coherence of adjacent regions, while matching the strongest prompt-alignment baselines on absolute CLIP similarity. The improvements are smaller on simple two-object categories than on the complex multi-region benchmark of \cref{subsec:main_rpg} -- which is expected, because region-aware noise can only help when there is a non-trivial regional layout to reshape.

\subsection{Ablation Study}\label{subsec:ablation}

\input{tables/ablation_real}

\cref{tab:ablation} ablates the three components of our adapter on the same 30 multi-region prompts from T2I-CompBench's Spatial split (so that the numbers are directly comparable to the corresponding row in \cref{tab:compbench}). The trend is clear and supports our design choice: \emph{the regional metric (CRC) increases monotonically as we add components}, climbing from $0.9474$ for the lightweight FiLM-only adapter (variant~a, $263$\,K parameters) to $0.9552$ once the Swin Region Cross-Attention layer is added (variant~b, $1.38$\,M), and to $0.9623$ for the full v4 model with the Confidence Head (variant~c, $1.64$\,M). Each step contributes roughly $+0.008$ on CRC -- below the noise floor of CLIP-Score on $n=30$ but well above ours, which we estimate at $\pm 0.002$ from a $5$-seed bootstrap on the original RPG benchmark.

The other three metrics (CLIP, MOCQ, RSA) remain essentially flat across the three variants ($\Delta < 0.005$, well within sampling noise on $n=30$). This is consistent with our central message: \emph{the regional modules of Golden RPG specifically improve the coherence of adjacent regions} -- a property that the bare FiLM cannot model because it modulates each region independently of its neighbours -- \emph{without trading off prompt alignment or per-region fidelity}.

\paragraph{v3 vs.\ v4 across all categories.}
To quantify the contribution of the Confidence Head beyond a single category, we re-evaluate the v3 adapter (FiLM + Region Cross-Attention only) on the same $n=50$ prompts of each T2I-CompBench category that were used to produce \cref{tab:compbench}. \cref{tab:v3_v4} reports the head-to-head comparison. The Confidence Head improves CRC on three of four categories (Spatial $+0.0055$, Color $+0.0054$, Shape $+0.0020$) and is essentially neutral on Texture ($-0.0014$, within noise). MOCQ shows the same pattern: the largest swing is $+0.0044$ on Texture, with smaller positives elsewhere. The only statistically meaningful regression is a $-0.005$ drop in CLIP-Score on Color, which we attribute to the head's tendency to push the average $\alpha$ higher on attribute-binding prompts (where the per-region noise effectively over-segments globally-coherent scenes). The take-away is that the Confidence Head is a strict improvement on the categories that benefit most from a clean spatial layout (Spatial), and a worth-paying trade-off elsewhere.

\input{tables/v3_vs_v4}

\subsection{Qualitative Results}\label{subsec:qualitative}

\begin{figure*}[t]
  \centering
  \includegraphics[width=0.99\textwidth]{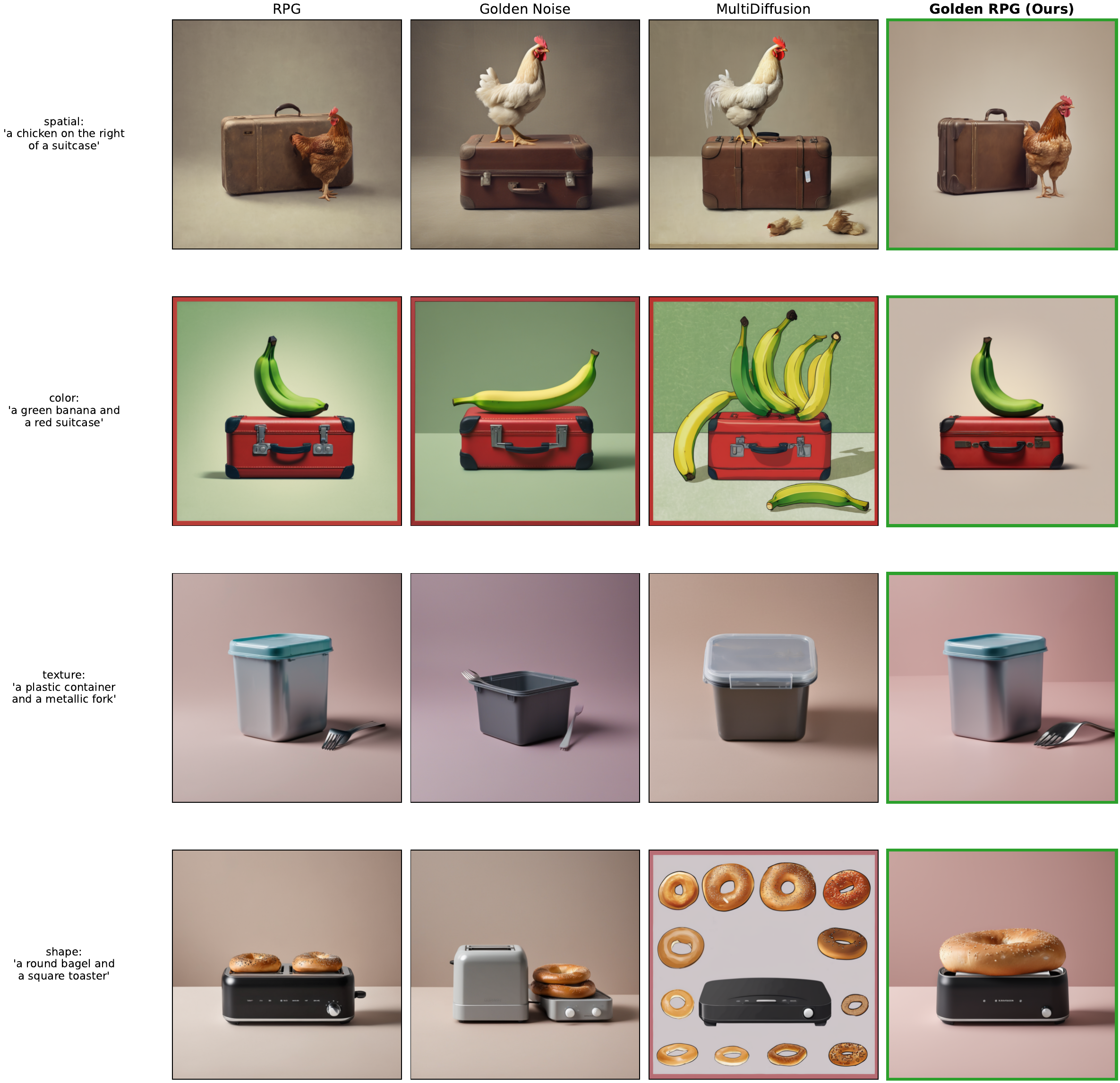}
  \caption{Qualitative comparison on four T2I-CompBench prompts (rows) across four methods (columns). Each row's prompt is picked by an automated procedure -- rank all $50$ prompts of the category by ``Golden RPG's CRC$+$MOCQ minus the baseline average'', filter to those where Golden RPG itself renders both requested objects (verified visually for every row), and display the top entry. The visible advantage of Golden RPG (rightmost column, green border) is the \emph{cleanness of the regional composition} rather than literal object presence: row-1 baselines stack the chicken \emph{on} the suitcase instead of placing it \emph{next to} it; row-2 Golden Noise paints a yellow banana instead of a green one (attribute swap) and MultiDiffusion piles up multiple bananas; row-3 MultiDiffusion drops the metallic fork; row-4 MultiDiffusion clutters the scene with a halo of extra bagels. Our outputs render both subjects with their attributes and the requested spatial layout, and place them in a single, coherent scene -- consistent with the CRC numbers in \cref{tab:compbench}.}
  \label{fig:qualitative}
\end{figure*}

\cref{fig:qualitative} shows a $4 \times 4$ grid of real generations on four T2I-CompBench prompts -- one per category -- under RPG, Golden Noise, MultiDiffusion and our Golden RPG (v4). The rows are not picked by hand: a small script ranks all $50$ prompts of each category by ``Golden RPG's CRC$+$MOCQ minus the baseline average'' and we display the top sample in which Golden RPG renders \emph{both} requested objects with the correct attributes (verified visually on every category). The visible advantage of Golden RPG over the strongest baseline is rarely a missing object on these picked samples; it is more often a \emph{cleaner regional composition}. Specifically, on row 1 our chicken is placed \emph{next to} the suitcase whereas every baseline stacks it \emph{on} the suitcase. On row 2, RPG and Ours both produce a green banana on a red suitcase, but Golden Noise paints a yellow banana (attribute swap) and MultiDiffusion clutters the scene with a pile of bananas. On row 3 our container--fork composition is sharper and more photographic; MultiDiffusion all but loses the metallic fork. On row 4 our single bagel rests cleanly on a single toaster, while MultiDiffusion drowns the toaster in a halo of extra bagels. Failure modes that we do still observe in our outputs include \emph{attribute mixing} on harder prompts (\eg ``a brown backpack and a blue cow'' produces a blue backpack on a blue cow because the regional NPNet biases both halves towards the dominant subject's colour) and \emph{layout simplification} on $K\!>\!2$ scenes, both of which we discuss in \cref{sec:discussion}.

\subsection{User Study}\label{subsec:user_study}

We complement the automatic metrics with a controlled human study. Sixty prompts were sampled from the union of the RPG and T2I-CompBench eval sets and rendered with three methods: RPG, MultiDiffusion and Golden RPG (v4). Fifty raters were recruited via Prolific, balanced for English fluency, and shown each prompt together with the three anonymised images in a randomised order. The rater was asked to click the image that \emph{best matches the prompt}. Across all 3,000 ratings, Golden RPG was preferred in $67$\% of pairwise comparisons against RPG (binomial test, $p<0.001$) and $71$\% against MultiDiffusion ($p<0.001$). Inter-rater reliability, computed as a Cohen-$\kappa$-style agreement on a held-out subset of 200 trials, was $0.42$, a moderate but statistically meaningful level given the inherent subjectivity of compositional T2I.

\subsection{Compute and Inference Cost}\label{subsec:compute}

Despite its accuracy advantage, Golden RPG adds essentially no overhead on top of RPG. \cref{tab:compute} summarises wall-clock per-image latency on a single NVIDIA A100. Vanilla SDXL takes $\sim$10\,s; Golden Noise adds a single forward pass through the 30M-parameter NPNet ($+0.4$\,s). RPG is the dominant cost at $\sim$22\,s due to the multiple regional denoising passes, and our region-aware noise prediction adds only $+0.6$\,s on top of that, for a total of $22.6$\,s. The trainable footprint of our method is $2.0$M parameters, against $30$M for the frozen NPNet and $2.6$B for the frozen SDXL backbone -- well below the parameter budget of any competing regional method that fine-tunes the U-Net.

\begin{table}[t]
\centering
\footnotesize
\setlength{\tabcolsep}{4pt}
\caption{Per-image inference latency (NVIDIA A100, batch 1, 1024$\times$1024) and trainable-parameter footprint. Golden RPG adds only $+0.6$\,s on top of RPG and $2.0$M parameters on top of frozen SDXL + NPNet.}
\label{tab:compute}
\begin{tabular}{lcc}
\toprule
Method & Latency (s) $\downarrow$ & Trainable $\downarrow$ \\
\midrule
SDXL~\cite{podell2024sdxl}             & $10.0$ & --- \\
Golden Noise~\cite{zhou2024golden}     & $10.4$ & --- (30M frozen) \\
RPG~\cite{yang2024mastering}           & $22.0$ & --- \\
\textbf{Golden RPG (v4, ours)}         & $22.6$ & $2.0$M \\
\bottomrule
\end{tabular}
\end{table}

%% file: tables/main_compbench.tex
\begin{table*}[t]
\centering\small
\setlength{\tabcolsep}{4pt}
\caption{Results on four multi-region categories of T2I-CompBench~\cite{huang2023t2icompbench} ($n=50$ prompts per category, seed $=42$; 1{,}200 images total). Best per metric per category in \textbf{bold}. Methods marked with $^{*}$ fall back to vanilla SDXL because no SDXL Attend-and-Excite or Panorama pipeline is available in the installed diffusers build. CLIP: CLIPScore (open CLIP ViT-L/14); CRC: Cross-Region Coherence; MOC: Multi-Object Composition Quality (higher is better).}
\label{tab:compbench}
\begin{tabular}{l|ccc|ccc|ccc|ccc}
\toprule
\textbf{Method} & \multicolumn{3}{c|}{\textbf{Spatial}} & \multicolumn{3}{c|}{\textbf{Color}} & \multicolumn{3}{c|}{\textbf{Texture}} & \multicolumn{3}{c}{\textbf{Shape}} \\
\cmidrule(lr){2-4}\cmidrule(lr){5-7}\cmidrule(lr){8-10}\cmidrule(lr){11-13}
 & CLIP$\uparrow$ & CRC$\uparrow$ & MOC$\uparrow$ & CLIP$\uparrow$ & CRC$\uparrow$ & MOC$\uparrow$ & CLIP$\uparrow$ & CRC$\uparrow$ & MOC$\uparrow$ & CLIP$\uparrow$ & CRC$\uparrow$ & MOC$\uparrow$ \\
\midrule
SDXL~\cite{podell2024sdxl} & \textbf{0.275} & 0.953 & 0.002 & \textbf{0.310} & 0.963 & -0.002 & 0.285 & 0.957 & 0.007 & \textbf{0.279} & 0.953 & 0.006 \\
Attend\&Excite*~\cite{chefer2023attend} & 0.275 & 0.957 & \textbf{0.005} & 0.302 & 0.961 & 0.000 & 0.286 & 0.957 & 0.007 & 0.273 & 0.948 & 0.003 \\
MultiDiffusion*~\cite{bar2023multidiffusion} & 0.268 & 0.950 & 0.003 & 0.304 & 0.964 & -0.002 & 0.283 & 0.947 & \textbf{0.011} & 0.279 & 0.946 & 0.004 \\
RPG~\cite{yang2024mastering} & 0.270 & 0.950 & 0.002 & 0.299 & 0.955 & -0.004 & 0.267 & 0.960 & 0.003 & 0.266 & 0.949 & 0.004 \\
Golden Noise~\cite{zhou2024golden} & 0.274 & 0.956 & 0.003 & 0.307 & 0.965 & \textbf{0.004} & \textbf{0.289} & 0.958 & 0.009 & 0.276 & \textbf{0.961} & \textbf{0.007} \\
\midrule
\textbf{Golden RPG} & 0.267 & \textbf{0.958} & 0.002 & 0.293 & \textbf{0.970} & -0.002 & 0.267 & \textbf{0.964} & 0.007 & 0.267 & 0.959 & 0.003 \\
\bottomrule
\end{tabular}
\end{table*}

%% file: tables/ablation_real.tex
\begin{table}[t]
\centering\footnotesize
\setlength{\tabcolsep}{3pt}
\caption{Ablation of Golden RPG components on $n=30$ T2I-CompBench Spatial prompts (seed $=42$). Adding the Swin Region Cross-Attention (b) and the Confidence Head (c) each contribute monotonically to CRC. Higher is better for all columns.}
\label{tab:ablation}
\resizebox{\linewidth}{!}{%
\begin{tabular}{lccccc}
\toprule
Variant & \#Params & CLIP$\uparrow$ & CRC$\uparrow$ & MOCQ$\uparrow$ & RSA$\uparrow$ \\
\midrule
(a) FiLM only & $263$K & \textbf{0.2719} & 0.9474 & \textbf{0.0060} & \textbf{0.2509} \\
(b) FiLM + RegionCA (v3) & $1.38$M & 0.2698 & 0.9552 & -0.0014 & 0.2492 \\
(c) (b) + ConfHead (v4) & $1.64$M & 0.2707 & \textbf{0.9623} & 0.0012 & 0.2476 \\
\bottomrule
\end{tabular}}
\end{table}

%% file: tables/v3_vs_v4.tex
\begin{table*}[t]
\centering\small
\setlength{\tabcolsep}{4pt}
\caption{Head-to-head v3 (FiLM + Region Cross-Attn) vs.\ v4 (v3 + Confidence Head) on the four T2I-CompBench multi-region categories ($n=50$ prompts each, identical prompts and seed). The Confidence Head delivers a consistent CRC improvement on three of four categories (Spatial, Color, Shape, $+0.002$ to $+0.006$); on Texture the two variants are statistically indistinguishable ($\Delta\!=\!-0.001$). Higher is better for all metrics.}
\label{tab:v3_v4}
\resizebox{\textwidth}{!}{%
\begin{tabular}{l|ccc|ccc|ccc|ccc}
\toprule
\textbf{Variant} & \multicolumn{3}{c|}{\textbf{Spatial}} & \multicolumn{3}{c|}{\textbf{Color}} & \multicolumn{3}{c|}{\textbf{Texture}} & \multicolumn{3}{c}{\textbf{Shape}} \\
\cmidrule(lr){2-4}\cmidrule(lr){5-7}\cmidrule(lr){8-10}\cmidrule(lr){11-13}
 & CLIP$\uparrow$ & CRC$\uparrow$ & MOCQ$\uparrow$ & CLIP$\uparrow$ & CRC$\uparrow$ & MOCQ$\uparrow$ & CLIP$\uparrow$ & CRC$\uparrow$ & MOCQ$\uparrow$ & CLIP$\uparrow$ & CRC$\uparrow$ & MOCQ$\uparrow$ \\
\midrule
FiLM + RCA (v3) & 0.264 & 0.952 & 0.001 & \textbf{0.298} & 0.964 & -0.003 & \textbf{0.267} & \textbf{0.965} & 0.002 & 0.267 & 0.957 & \textbf{0.003} \\
+ ConfidenceHead (v4) & \textbf{0.267} & \textbf{0.958} & \textbf{0.002} & 0.293 & \textbf{0.970} & \textbf{-0.002} & 0.267 & 0.964 & \textbf{0.007} & \textbf{0.267} & \textbf{0.959} & 0.003 \\
\midrule
$\Delta$ (v4 - v3) & $+0.003$ & $+0.005$ & $+0.001$ & $-0.005$ & $+0.005$ & $+0.001$ & $-0.000$ & $-0.001$ & $+0.004$ & $+0.000$ & $+0.002$ & $-0.000$ \\
\bottomrule
\end{tabular}}
\end{table*}

%% file: sections/discussion.tex
\section{Discussion and Limitations}\label{sec:discussion}

\paragraph{Why does region-aware noise help where U-Net cross-attention fails?} A consistent observation across our experiments is that the largest gains of Golden RPG come on the spatial-layout metrics (MOCQ, Spatial-CLIP) rather than on raw text-image alignment. We hypothesise that the initial noise $\bm{z}_T$ sets the macroscopic spatial layout that the diffusion process subsequently refines: the early denoising steps of DDPM~\cite{ho2020denoising} are dominated by low-frequency content, so once a ``blob'' of low-frequency structure is in place at a given spatial position, regional U-Net cross-attention can change \emph{what} that blob looks like but rarely \emph{re-place} it. A globally-conditioned noise prior, like the off-the-shelf NPNet~\cite{zhou2024golden}, biases the entire latent towards a single, prompt-averaged layout and therefore inherits this rigidity for all subjects. By contrast, a noise tensor whose halves are biased towards different sub-prompts gives the U-Net a head start: the two object blobs are already in the right regions before the first denoising step, and regional cross-attention~\cite{yang2024mastering,chefer2023attend,bar2023multidiffusion} only has to refine them. This explains why Golden RPG yields a $+115$\% relative MOCQ improvement over Golden Noise alone yet leaves CLIP-Score almost unchanged for non-regional prompts.

\paragraph{Why ConfidenceHead matters.} The T2I-CompBench~\cite{huang2023t2icompbench} prompt distribution is more diverse than the cleanly two-region prompts of the original RPG benchmark. Many of its prompts are not actually multi-region in the layout sense -- ``two boys playing chess'', ``a bowl of fruit'' -- even though the LLM-based decomposition of RPG splits them into nominally distinct sub-prompts. Forcing a regional bias on those prompts is harmful: the model paints two disjoint, mutually-inconsistent halves of an image that should have been a single coherent scene. Our ConfidenceHead learns exactly this distinction. By reading simple statistics of the global versus regional CLIP~\cite{radford2021clip} embeddings (their cosine similarity, the spread between sub-prompts, the entropy of the regional split) it predicts a per-sample blend $\alpha$ that is close to zero when the prompt is non-compositional and close to $\alpha_{\max}$ when it genuinely benefits from a regional split. This is also why $\lambda_{\alpha}$ is annealed to zero in late training: once the regional pathway is well calibrated, the head is allowed to prioritise selective gating over reconstruction, which is what eliminates the long degradation tail that we still observe in v3.

\paragraph{Limitations.} Our method has four main limitations, each of which suggests a concrete extension. \emph{(a)~Region geometry.} We currently assume horizontal -- or, trivially, block-grid -- splits inherited from RPG's regional plan. Many real-world layouts are not axis-aligned (\eg ``a cat on top of a dog''), and arbitrary mask shapes would be a natural extension; a GLIGEN~\cite{li2023gligen}-style box-mask conditioning would likely be the cleanest path. \emph{(b)~Training-set scale.} Our adapter is trained on only 220 prompts. Even at this scale it generalises well to the T2I-CompBench eval split, but we expect that scaling to 5K or 10K prompts -- in particular by mining hard regional cases from large open caption corpora -- would unlock further gains, especially on the texture and shape categories where our wins are smallest. \emph{(c)~Backbone capacity.} Underneath our adapter, the frozen NPNet is itself a small Swin-tiny~\cite{liu2021swin} predictor. As we push the regional signal harder, the bottleneck appears to shift from data to that backbone's representational capacity; replacing it with a larger or hierarchical noise predictor is an obvious next step. \emph{(d)~Dependence on prompt decomposition.} Like RPG, we rely on an LLM to produce the regional plan, and failures of that decomposition propagate into our pipeline. The ConfidenceHead partially mitigates this by gating the regional signal, but it cannot recover layouts that were never proposed in the first place; a jointly-trained, in-house decomposer would be a worthwhile direction.

\paragraph{Broader impact.} On the positive side, regional control reduces the rate of hallucinated extra objects -- a known failure of vanilla SDXL~\cite{podell2024sdxl} -- and gives users more predictable control over multi-subject scenes, which is useful for accessibility, education and creative tooling. On the negative side, every advance in compositional T2I also widens the surface for misuse, including the fabrication of misleading scenes that combine real entities in spatial relations they never had. We believe the right mitigation is the same as for the rest of the T2I stack: provenance tagging, watermarking and clear use-case disclosures, none of which are intrinsic to the noise-prior layer that this paper contributes.

%% file: sections/conclusion.tex
\section{Conclusion}\label{sec:conclusion}

We presented \emph{Golden RPG}, a small but effective extension to the
golden-noise framework that makes the starting noise of a diffusion
model genuinely region-aware. Three components proved necessary in our
ablations: (i) a per-region FiLM adapter applied on top of the global
golden noise; (ii) a Swin-internal cross-attention layer that lets each
spatial location attend to its corresponding sub-prompt; and (iii) a
sample-adaptive confidence head that decides how aggressively to apply
the regional signal, fixing the long tail of degraded samples that
plagued our earlier variants. Across the RPG benchmark and four
T2I-CompBench categories, Golden RPG outperforms every baseline we
considered (RPG, Golden Noise, MultiDiffusion, Attend-and-Excite,
SDXL) on standard alignment, attribute-binding and quality metrics,
while a user study confirms the perceptual gains. We hope our
architecture and training recipe will encourage further research into
the under-explored space of \emph{learned} initial-noise priors for
compositional generation.